\documentclass{article}

\PassOptionsToPackage{square,sort,comma,numbers}{natbib}

    \usepackage[preprint]{neurips_2020}

\usepackage[utf8]{inputenc} 
\usepackage[T1]{fontenc}    
\usepackage{hyperref}       
\usepackage{url}            
\usepackage{booktabs}       
\usepackage{amsfonts}       
\usepackage{nicefrac}       
\usepackage{microtype}      
\usepackage{graphicx}
\usepackage{amsmath,amssymb}
\usepackage{algorithm, algorithmic}
\usepackage{wrapfig}
\title{SPACE: Structured Compression and Sharing of Representational Space for Continual Learning}

\author{Gobinda Saha\thanks{Authors contributed equally to this work} , Isha Garg$^*$, Aayush Ankit, Kaushik Roy\\
Department of ECE\\
Purdue University\\
West Lafayette, IN 47906, USA\\
{\tt\{gsaha,gargi,aankit,kaushik\}@purdue.edu}
}

\begin{document}

\maketitle
\vspace{-20pt}
\begin{abstract}
    Humans learn adaptively and efficiently throughout their lives. However, incrementally learning tasks causes artificial neural networks to overwrite relevant information learned about older tasks, resulting in `Catastrophic Forgetting'. Efforts to overcome this phenomenon often utilize resources poorly, for instance, by growing the network architecture or needing to save parametric importance scores, or violate data privacy between tasks. To tackle this, we propose SPACE, an algorithm that enables a network to learn continually and efficiently by partitioning the learnt space into a \textit{Core} space, that serves as the condensed knowledge base over previously learned tasks, and a \textit{Residual} space, which is akin to a scratch space for learning the current task. After learning each task, the Residual is analyzed for redundancy, both within itself and with the learnt Core space. A minimal number of extra dimensions required to explain the current task are added to the Core space and the remaining Residual is freed up for learning the next task. We evaluate our algorithm on P-MNIST, CIFAR and a sequence of 8 different datasets, and achieve comparable accuracy to the state-of-the-art methods while overcoming catastrophic forgetting. Additionally, our algorithm is well suited for practical use. The partitioning algorithm analyzes all layers in one shot, ensuring scalability to deeper networks. Moreover, the analysis of dimensions translates to filter-level sparsity, and the structured nature of the resulting architecture gives us up to $5$x improvement in energy efficiency during task inference over the current state-of-the-art.\footnote{Source code available at : \href{https://github.com/sahagobinda/CL\_PCA}{https://github.com/sahagobinda/CL\_PCA}}
\end{abstract}

\section{Introduction}\label{intro}
Deep learning has recently gained tremendous success in achieving human-level performance and surpassed human experts in a variety of domain-specific tasks \cite{resnet, algo}. However, in the context of continual learning (CL) \cite{cla,clb,rev,cl3,owm}, whereas humans can continuously adapt and learn new tasks throughout their lifetimes without forgetting the old ones, current deep learning methods struggle to perform well since the data distribution changes over the course of learning. This problem arises in Artificial Neural Networks (ANNs) because of the way input data is mapped into the network's parametric representation during learning. When the input data distribution changes, network parameters are updated through gradient-based methods to minimize the objective function with respect to the data of the current task only, without heed to the distribution learned over the previous tasks. This leads to a phenomenon called `\textit{Catastrophic Forgetting}'~\cite{cat1,cat2}, wherein the network forgets how to solve the older tasks upon exposure to new ones.  

Efforts to overcome catastrophic forgetting can be divided into three prominent groups. The first set of methods involves a form of rehearsal over older tasks while training the current task, such as in \cite{robbo,icarl,gem,er,dgr}. These methods may prove infeasible in the real-world setting, where we have limited memory budgets and access to older data due to privacy constraints. The second line of research focuses on regularization to penalize changes in important parameters, as in \cite{ewc,si,mas,cl4}. These methods make more efficient use of the available resources, but suffer from catastrophic forgetting as the number of tasks increase. The third set of techniques, which our algorithm falls under, works on the concept of parametric isolation \cite{pisol}, where disjoint or overlapping sets of parameters are allocated to separate tasks, such as in \cite{packnet,pathnet}. Some methods under this category like \cite{den,progress} grow the network with subsequent tasks, but are impractical in resource constrained environments or result in unstructured sparsity that cannot be utilized well in general purpose hardware.

To address the problem of catastrophic forgetting in an energy, memory and data constrained environment, we propose SPACE, a Principle Component Analysis (PCA) \cite{pca} based approach to identify redundancy within the filters (or neurons in fully-connected layer) for a given task. We frame this in terms of finding the significant dimensionality of the internal space of representations generated by the layer's filters. We let the network learn the relevant filters that adhere to this reduced dimensionality, and freeze them while learning the next task to avoid forgetting.
The concept of successively pruning and/or freezing parameters in a continual learning setup has been explored in many previous works such as in \cite{progress}, PackNet \cite{packnet} and HAT \cite{hat}. PackNet and HAT hold state-of-the-art (SOTA) performance in the experimental setup considered in this paper, and are conceptually similar to our work. However, there are some key distinctions. PackNet pre-assigns a fixed sparsity level for all tasks and layers irrespective of task complexity, sufficiency of learned features at a layer, or available resources. To avoid fixed sparsity levels, HAT learns almost-binary masks over feature maps per task using gated attention. In contrast, our algorithm allocates network resources (filters or neurons) in each layer for each task through variance-based statistical analysis of network representations (pre-ReLU activations). Such analysis explicitly takes into account the complexity of a given task and its overlap with the learned representations of the past tasks. In doing so, our method identifies task-specific subspaces of representations which, in conjunction with past representations, are sufficient to attain good performance on the current task. An important consequence of this formulation is that we do not ascribe any importance to individual, static network elements. Instead, we only identify the dimensionality of the relevant subspaces and let standard learning methods learn a new set of relevant parameters. Additionally, unlike PackNet, our method creates structured sparsity in the network which can be leveraged well in general purpose hardware. We test our method on Permuted MNIST, split CIFAR-10/100, and a sequence of 8 datasets and compare our results with other relevant methods. We outperform Elastic Weight Consolidation (EWC) \cite{ewc}, Learning without Forgetting (LwF) \cite{lwf} and HAT in terms of accuracy, memory utilization and energy consumption. Since PackNet prunes at a finer grained sparsity level than us, we perform comparably or marginally lower in accuracy but the resulting structured sparsity enables us to be more energy efficient during inference over tasks. Moreover, we demonstrate scalability to deeper network architectures, while ensuring data privacy between tasks.
\section{Related Works}
There are three prominent lines of research to enable Deep Neural Networks (DNNs) to learn continually, mentioned briefly in section \ref{intro}. Here, we go into the details of the representative works and highlight their contributions and differences with our work.

The first line of research uses data augmentation in order to rehearse the data from the older tasks while training on the current task. This allows the network to jointly optimize the training distributions and alleviates the problem of catastrophic forgetting. In \cite{icarl},~\cite{gem}, and \cite{mega}, a subset of examples from the previous tasks are stored in the memory, and in \cite{dgr}, a generator is trained to create synthetic training data for the previous tasks. However, these techniques require extra memory to store samples or the parameters of the generator, which additionally violates data privacy. In contrast, our method does not need to store data, avoiding the need for extra memory and maintaining data privacy. 

The second set of methods overcome catastrophic forgetting by sharing weights among the tasks. Typically, a regularizer term adds a penalty to the loss function to discourage significant change in important parameters for the older tasks while learning new tasks. Kirkpatrick \textit{et al.}~\cite{ewc} introduce EWC (later modified by~\cite{oewc} for efficiency), which calculates the task specific parametric importance from the diagonal elements of the Fisher Matrix. Zenke \textit{et al.}~\cite{si} calculate importance based on the sensitivity of the loss with respect to the parameters. Li and Hoiem \textit{et al.}~\cite{lwf} introduce LwF, which adds a proxy distillation loss to the objective function to retain the activations of the initial network while learning new tasks. Ebrahimi \textit{et al.}~\cite{ucb} propose UCB, which regularizes weights of the  Bayesian neural networks based on uncertainty measures. These methods show varying degrees of success in overcoming catastrophic forgetting. However, the balance between rigidity and plasticity ultimately breaks down as the complexity and the number of tasks increase, resulting in performance degradation on the older tasks. In contrast, learning a reduced set of filters via SPACE allows us to make better use of resources, and freezing them alleviates catastrophic forgetting. 
   
The third set of techniques isolates task-specific parameters. This includes methods that adjust the architecture with pruning (PackNet) \cite{packnet}, growing \cite{progress, cpg, acl, apd, cl2}, activation masking (HAT)~\cite{hat} or introducing parallel paths \cite{pathnet,rpsnet} for new tasks. Rusu \textit{et al.}~\cite{progress} protect old knowledge by freezing the base model and adding new sub-networks with lateral connections for each new tasks. However, this method has limited scalability, as the network size increases significantly with number of tasks. Another representative work, which is the most relevant to our method PackNet. Leveraging the fact that DNNs are overparameterized \cite{overparam}, PackNet employs weight magnitude based pruning \cite{magprune,dsd} to free up unimportant parameters, and uses them for learning future tasks while keeping the old weights frozen. 
Our main distinction from PackNet is that our method utilizes structured sparsity \cite{sp}. Such sparsity can be leveraged better in hardware as shown by \cite{transformer}, and translates to better energy efficiency during inference.  Additionally, PackNet pre-assigns a heuristically chosen compression ratio for each layer and each task. Like our method, HAT eliminates the need for such pre-assigned sparsity levels. It does so by learning near-binary masks over feature maps, resulting in structured sparsity over parameters. The optimization objective in HAT encourages masks to be sparse, and the degree of sparsity and forgetting is controlled by two hyperparameters which are optimized via grid search. In contrast to both these methods, SPACE controls sparsity by analyzing the learned representations with an explainable variance threshold hyperparameter. This formulation allows varying levels of filter allocation per layer per task, enabling resource minimality by accounting for task complexity, shareability of features at different layers, and overlap of representations learned for current task with previous tasks. 
\vspace{-5pt}
\section{Method}\label{sec3}
\vspace{-5pt}
Here, we describe our method for incremental training of DNNs in the continual learning paradigm, wherein an unknown number of tasks with varying data distributions arrive at the model sequentially. 
\subsection{Overview}\label{sec0}
Conceptually, our algorithm aims to incrementally learn a minimally growing set of filters that contain the condensed knowledge of tasks seen so far. To achieve this, we partition filters in each layer into a set of \textit{core filters} and \textit{residual filters}. The activations generated when inputs convolve with these filters make up the core representational space and the residual representational space respectively. Henceforth, we drop the words `representational space' and just refer to them as Core and Residual. We add compressed information pertaining to the tasks previously seen to the Core filters, and the Residual filters are reserved for the current task only. Each new task is learned by following a three-step procedure: \textbf{First}, the network is trained on the current task. Both the residual and core filters are used to generate activations. All the information present in the core filters is leveraged but not updated, whereas the residual filters are modified freely. We achieve zero forgetting by freezing these core filters. Once the network converges on the current step, we decouple the residual and the core in the \textbf{second} step. We use the \textbf{Projection-Subtraction} algorithm (explained in section \ref{sec2}) to identify and remove the redundancies between the newly learned Residual and the frozen Core. This reduced redundancy, decoupled Residual is analyzed in the \textbf{third} step using PCA to identify redundancies within itself. We then determine the number of filters to be added from the Residual to the Core. The redundant filters are pruned-out and fine-tuning (retraining) is performed on these newly added filters to mitigate any accuracy drop. Filter selection and pruning is done concurrently for all layers, and hence we only retrain the network once per task. For the next task, the Core filter set has grown, and the smaller Residual set is freed up to learn the next task.

\subsection{Filter Redundancy Detection and Compression for a Single Task} \label{sec1}
Most common DNN architectures \cite{alex,vgg} have been shown to have highly correlated filters within each layer that potentially detect the same features. Garg \textit{et al.}~\cite{isha} find such correlations by applying PCA on the activation maps generated by these filters, and optimize the architecture by removing redundant filters and layers from a trained network. They trained this structurally compressed architecture from scratch, with negligible loss of accuracy. In this subsection we explain how this method is applied to learn a standalone task.

\begin{figure}
\begin{centering}
  \includegraphics[width=0.7\linewidth]{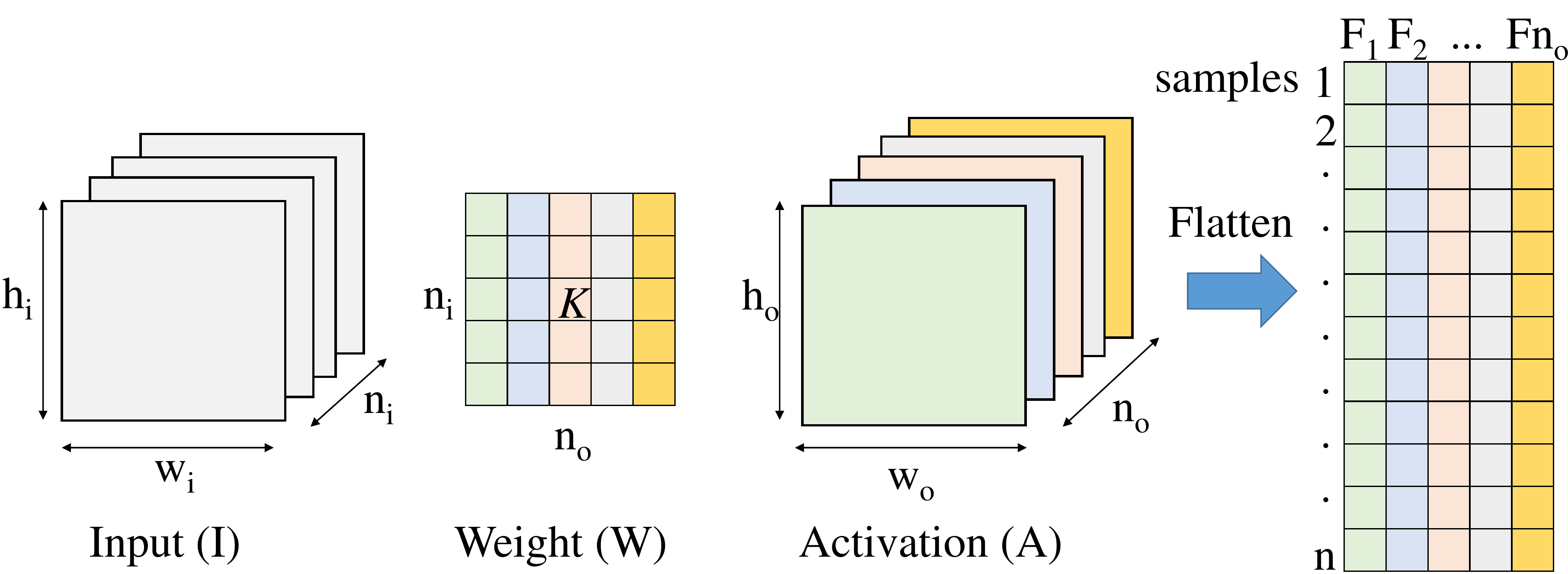}
  \caption{Illustration of the activation matrix generation for PCA from a typical convolutional layer. }
\label{fig:filt_red}
\end{centering}
\end{figure}

\textbf{Network Compression using PCA}. PCA \cite{pca} is a well-known dimensionality reduction technique used to remove redundancy between correlated features in a dataset. We are interested in detecting redundancies between filters and hence, we use the (pre-ReLU) activations, which are instances of filter activity, as feature values. Figure~\ref{fig:filt_red} illustrates the process of data collection for PCA. Let, $n_{i}$ ($n_{o}$) denote the number of input (output) channels for a convolutional layer and $h_{i},w_{i}$ ($h_{o},w_{o}$) denote the height and width of the input (output) feature maps. The 4D weight tensor for a particular layer is flattened to a 2D weight matrix ($W$) for illustration in the figure, where each filter, $F_{j}\in \mathbb{R}^{n_{i}\times k\times k}; j =1,2...,n_o $ is represented as a column of $W$, where $k$ is the kernel size. This filter acts upon a similar sized input patch from the feature map feeding into the layer. Let the input patch be, $I_{N} \in \mathbb{R}^N \text{ where } N={n_{i}\times k\times k} $, then the weight matrix can be represented as $W \in \mathbb{R}^{N\times n_{o}}$. Upon convolution with all the filters ($dot(I_{N},W)$), one input patch is represented as a vector in the internal representational space, $\mathbb{R}^{n_{o}}$. Since there are $h_{o}\times w_{o}$ input patches per example, for $m$ input examples there will be $n=m\times h_{o}\times w_{o}$ patches. Thus, after convolution with $m$ examples we obtain a flattened activity matrix $A \in \mathbb{R}^{n\times n_{o}}$, where $n$ represents the total number of samples. Since each row of this matrix corresponds to the internal representation of an input patch in $\mathbb{R}^{n_{0}}$ and each column represents instances of filter activity for all input patches over all examples, this 2D matrix is ideal for performing PCA and detecting redundancy in filter representations. 

Filters are correlated if their convolution with different input patches across many samples produces similar patterns of activations. The rank of $A$ can then be reduced without significant loss in accuracy. This means that the internal representations lie in a lower dimensional subspace, $\mathbb{R}^{p}$ inside $\mathbb{R}^{n_{o}}$, where $p<n_{o}$. Applying PCA on mean-normalized $A$, we find the number of principal components, $p$ required to cumulatively explain $x\%$ of the total variance in the input, where the variance threshold ($x$) is usually chosen in the range of 90 to 99.9. Garg \textit{et al.}~\cite{isha} use this as the number of filters needed at each layer to learn efficient representations, found concurrently at all layers. The redundant filters at each layer are zeroed out, and the network with the optimized architecture is retrained on the current task to gain back any drop in accuracy.

\subsection{Sharing Representational Space for Incrementally Learned Tasks}\label{sec2}
In the context of continual learning, we learn the first task in a compact manner by the approach described in the previous section. Since for the first task there are no core filters, the algorithm treats the entire representational space as the residual, and selects the number of filters that capture $x\%$ of the total variances in this space. While this algorithm is effective in finding a reduced dimensionality architecture for a single task, it suffers from catastrophic forgetting in sequential task learning scenarios. PCA analyzes the entire activation space for low dimensional representations of only the current data, and cannot accommodate preservation of spaces relevant to older tasks (Core space). Therefore for learning new sequential tasks, we freeze the core filters and modify the residual filters. We determine the number of new filters needed for the current task by jointly analyzing the Core and Residual spaces. To effectively do this, we first decouple the Residual from the Core, and analyze this decoupled Residual for redundancy. By this formulation, we share the core filters among the tasks and add new task specific filters to the Core from Residual as needed for each new task.

\begin{figure}
\begin{centering}
  \includegraphics[width=1\linewidth]{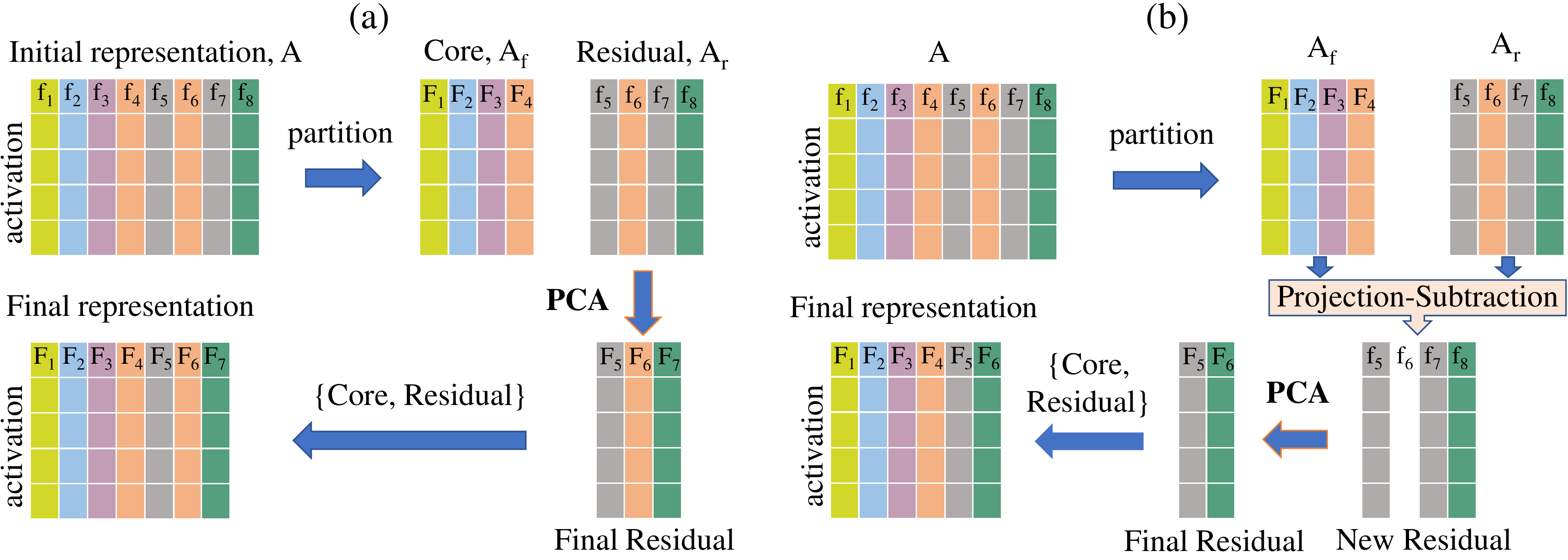}
  \caption{Decoupling Core and Residual for redundancy detection. Similar colors represent correlation. In (a), we perform PCA on the Residual directly, which fails to detect a redundancy in F4 and F6. In (b), before performing PCA, we subtract from the Residual its projection on the Core, thus removing the duplicacy.}
\label{fig:explain}
\end{centering}
\vspace{-10pt}
\end{figure} 

\textbf{Core and Residual Spaces.} Let there be $f$ frozen and $r$ trainable filters in a layer for a new task, such that number of original filters is $n_o = f+r$. The $r$ filters are randomly initialized after being pruned in the previous task. First, the network is trained until convergence on the current task, updating the trainable filters only (the frozen filters are still used, just not updated). Then, the activation matrix ($A$) is generated from the data of the current task.  
To allow for preservation of filters relevant to older tasks, we mean normalize and partition $A$ into two sub-matrices \(A=\begin{bmatrix}A_f & A_r\end{bmatrix}\) such that $A_f \in \mathbb{R}^{n\times f}$ and $A_r \in \mathbb{R}^{n\times r}$. The total variance in $A$ is $v_T=v_f+v_r$, where $v_f(=\frac{1}{n-1}\text{tr}(A_f^TA_f))$ and $v_r(=\frac{1}{n-1}\text{tr}(A_r^TA_r))$ are core and residual variances respectively. Since $f$ core filters carry the knowledge from the old tasks, we freeze them while learning any future tasks. Our algorithm determines how many new filters from the residual need to be added to the core to explain $x\%$ of the total residual variance ($v_r$). One possibility to determine this number is to apply PCA directly on $A_r$. However, when learning $A_r$, we do not explicitly encourage decorrelation between the residual and the core spaces. This means that there is a possibility that we might learn residual filters that produce activation patterns that are correlated to the core activations, leading to redundancy between the core and residual spaces. Figure \ref{fig:explain} (a) shows one such scenario where we might overestimate the number of filters as a consequence of such duplication.

\textbf{Decoupling Core and Residual via Projection-Subtraction.} To overcome this, we construct a new residual activation matrix, $A_{r}^{new}$ from $A_r$ by removing the activations that are correlated to the core activations by applying Projection-Subtraction step. In the \textbf{Projection} step, we construct $A_r^{proj}$ by projecting $A_r$ onto the space described by the (column) basis of $A_f$. We obtain these bases by reduced Singular Value Decomposition (SVD) \cite{stang} of $A_f$ : $A_f=U_f\Sigma_f V_f^T$, where columns of $U_f\in \mathbb{R}^{n\times f}$ are the desired bases. Then we perform projection by :
\begin{equation}\label{eq:eq1}
    A_r^{proj}= (U_fU_f^T)A_r.
\end{equation}
With this formulation, if there was complete duplicacy at a task, and the core filters sufficiently describe all the variance in the residual, then $A_r^{proj}$ would be exactly $A_r$. In this case, we do not need to add any new filters to the core. Let $v_r^{proj}$ denote the amount of variance in residual activations that is explained by the core.
\begin{equation}
    v_r^{proj} =\frac{1}{n-1}\text{tr}((A_r^{proj})^TA_r^{proj})
\end{equation}

This means that the $f$ frozen filters for the current task capture not just the core variance, $v_f$ but also explain a fraction $(v_r^{proj}/v_r)$ of the residual variance. Supplementary section \ref{S1.1} outlines the projection methodology in more detail. 
In the \textbf{Subtraction} step we construct a matrix $A_r^{new}$, which captures the residual activation patterns that are not explained by the core, or the variance in the representation unique to the new task.    
\begin{equation}\label{eq:eq2}
    A_r^{new}=A_r-A_r^{proj}
\end{equation}

Finally, by applying PCA on  $A_r^{new}$, we determine how many new filters from the residual are cumulatively required to explain $x\%$ of residual variance in conjunction with the $f$ frozen filters. Figure \ref{fig:explain}(b) illustrates how Projection-Subtraction-PCA steps ensures minimal filter addition by avoiding filter duplication.

\textbf{One-Shot Pruning and Retraining.} Let's say performing PCA after Projection-Subtraction identifies the need for $l$ new filters at a layer, where $0\leq l<r$. We prune $(f+l+1)^{th}$ to $n_0^{th}$ filters in that layer. These newly identified $l$ filters are retrained with frozen $f$ filters to mitigate any drop in accuracy for the current task. The activation matrices for PCA for all layers are collected concurrently and independently for all layers by just forward passes, similar to \cite{isha}. This allows us to prune all layers simultaneously, in one-shot. Consequently, we require only one retraining step per task to mitigate any drop in accuracy. The pseudo-code of the algorithm is given in Supplementary section \ref{S1.2}.

\section{Experimental Setup}
\textbf{Datasets.} We evaluate our algorithm on P-MNIST \cite{mnist}, and the spilt versions of CIFAR-10 and CIFAR-100 datasets \cite{cifar}. To test the performance on inter-dataset tasks, we use the compilation of 8-datasets introduced in \cite{hat}. The details of this dataset are given in supplementary section \ref{S2.1}. The P-MNIST dataset is created out of the MNIST dataset by randomly permuting all the pixels of the MNIST images differently for $10$ tasks, where each task has $10$ classes. The Split versions of CIFAR-10 and CIFAR-100 datasets are constructed by splitting the original datasets into $5$ and $10$ sequential tasks, with $2$ and $10$ classes per task, respectively. 

\textbf{Network Architecture.} For P-MNIST, we use a multi-layered perceptron (MLP) with two hidden layers with $1000$ neurons each, and ReLU activations. Though we described our method for a typical convolutional layer in section \ref{sec3}, it is readily applicable to fully-connected layers. We construct the activation matrix, $A\in\mathbb{R}^{m\times n}$ such that $m$ is the number of input examples and $n$ is the number of neurons. For CIFAR-10 and CIFAR-100, we use networks with five convolutional layers followed by a classifier. For the CIFAR-10 experiments, we use $32-32-64-64-128$ filters in the five convolutional layers whereas for CIFAR-100 we double them to $64-64-128-128-256$ filters. To show the scalability of our method to deeper networks, we test the 8-dataset tasks on a reduced ResNet-18, similar to~\cite{gem} with base number of filters as 40. Supplementary section \ref{S2.1} contains details on the architecture and training settings.


\textbf{Selection of Variance Threshold.} We treat task-wise variance threshold ($x$) as a hyperparameter, with some guidelines that make it easy to optimize. For a given task, higher values of $x$ yield higher classification accuracy, at the expense of higher filter utilization. Continual learning in memory constrained (fixed resource) environments demands a trade-off between classification accuracy and parameter utilization per task. Tuning $x$ at a layer translates to a more explainable number of filters than the commonly used heuristic threshold selection choices. These choices include unintuitive pre-assigned fixed sparsity, or tunable sparsity that requires extensive grid search to optimize. Variance tuning also takes into account both task complexity and the shareability of features learned at a layer. Empirically, we find that instead of using uniform $x$ across all layers, setting higher variance threshold for the first layer always yields better classification accuracy. This is discussed in more detail in Supplementary section \ref{S2.2}, along with the chosen values of $x$.

\textbf{Baselines.} To establish a strong baseline, we consider Single Task Learning \textbf{(STL)}, where the full network with all resources is trained for each task separately. While this is not truly a continual learning setting, it can be treated as stacking separate, full sized networks for each task with no issue of catastrophic forgetting. To ensure a fair comparison, we  compared our algorithm with the most relevant methods that  do not grow the network architecture and do not store old data. Among these methods we consider Learning without Forgetting \textbf{(LwF)} and Elastic weight Consolidation \textbf{(EWC)}, which use activation and weight-specific regularizers, respectively, and utilize the entire network. In addition, we compare with SOTA methods \textbf{HAT} and \textbf{PackNet} which isolate important parameters though masking.~The implementation details for all these methods are given in Supplementary section \ref{S2.4}. Additionally, comparison with \textbf{UCB} is provided in Supplementary section \ref{here}. 

\vspace{-5pt}
\section{Results and Discussion}\label{sec5}
\vspace{-5pt}

We evaluate the performance of our algorithms in terms of classification accuracy, utilization of network resources and inference energy. Classification accuracy is cumulative. The reported task accuracy is evaluated over all tasks seen up to the current task, inclusively.

\begin{figure}[t]
\begin{center}
  \includegraphics[width=1\linewidth]{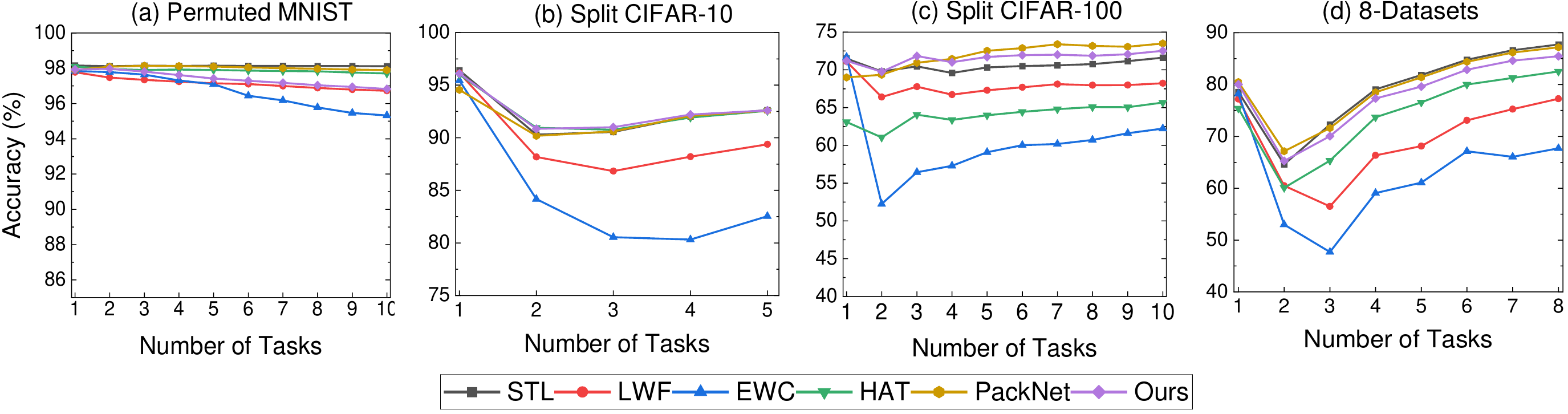}
\end{center}
  \caption{Classification accuracy over incrementally learned tasks for different datasets.}
\label{fig:long}
\label{fig:accu}
\end{figure}


\begin{table*}[t] 
\centering 
\scalebox{0.85}{
\begin{tabular}{l l c c c c c c} 
\toprule 
\cmidrule(l){2-8} 
\textbf{Dataset}& Metric & STL$^*$ & LwF\cite{lwf} & EWC\cite{ewc} & HAT\cite{hat} & PackNet\cite{packnet} & SPACE(Ours)\\ 
\midrule 
 & Accuracy (\%) & 98.11\% & 96.72\% & 95.32\% & 97.63\% & \textbf{97.89\%} & 96.82\% \\ 
P-MNIST&Network Size & 10 & 1 & 1 & 0.74 & 0.61 & \textbf{0.58}\\ 

&Inference Energy &1	&1.04&	1.05&	0.64 & 0.96&	\textbf{0.62}\\

\midrule 
 & Accuracy (\%) & 92.63\% & 89.38\% & 82.54\% & 92.56\% & 92.58\% & \textbf{92.60\%} \\ 
CIFAR-10&Network Size & 5 & 1 & 1 & 0.80 & \textbf{0.70} & \textbf{0.70} \\ 
&Inference Energy  &1	&1.02	&1.02 & 0.89	&0.99&	\textbf{0.68}\\

\midrule 
 & Accuracy (\%)& 71.62\% & 68.23\% & 62.22\% &65.67\% & \textbf{73.49\%} & 72.53\% \\ 
CIFAR-100&Network Size & 10 & 1 & 1 & 0.83& 0.80 & \textbf{0.79}\\
&Inference Energy &1	&0.97&	0.99& 0.75&0.97&	\textbf{0.56}\\


\midrule 
& Accuracy (\%)& 87.13\% & 77.28\% & 67.69\% &82.53\% & \textbf{86.31\%} & 85.48\% \\ 
8-Datasets &Network Size & 8 & 1 & 1 & 0.95& \textbf{0.90} & 0.93\\
&Inference Energy &1	&1.01&	1.01&	0.81 &1&	\textbf{0.80}\\
\bottomrule 
\end{tabular}}
\caption{Comparison of classification accuracy (\%), network size (relative to original network) and average inference energy per task (normalized with respect to STL, the lower the better). *STL is not a CL method.} 
\label{tab:tab1} 
\vspace{-5pt}
\end{table*}

\textbf{Accuracy and Network Size}. The incremental classification accuracy is plotted against number of tasks learnt in Figure~\ref{fig:accu}. The final classification accuracy and network size after training all the tasks for different methods and datasets are enumerated in Table~\ref{tab:tab1}. Note that STL is not a CL method and uses a separate network per task. For all datasets, SPACE consistently outperform regularization based methods LwF and EWC significantly in accuracy while utilizing considerably lesser resources. For P-MNIST, while STL, HAT and PackNet are marginally more accurate, our method achieves the highest compression ratio, using only $58\%$ of the original network parameters. Figures \ref{fig:accu}(b) and (c) show results for the split CIFAR-10 and CIFAR-100 datasets. Table~\ref{tab:tab1} shows that SPACE performs comparably to STL on CIFAR-10 and is $\sim1\%$ better on CIFAR-100 (presumably due to regularization) while using $7.1$x and $12.7$x fewer parameters, respectively. Notably, we have marginally better accuracy than both HAT and PackNet on CIFAR-10, with comparable or lesser parameters. On CIFAR-100 we are $\sim7$\% more accurate than HAT with lesser parameters, but are marginally lesser accurate than PackNet.
We next consider a challenging setup, by scaling to a deeper architecture (ResNet18) and learning inter-dataset tasks. On the 8-dataset setup, SPACE performs better than all considered CL baselines except PackNet, which benefits from the finer granularity of pruning employed. Thus, we show that SPACE is an accurate and efficient algorithm for large-scale real world CL applications.

\begin{figure}[t]
    \centering
    \begin{minipage}{0.47\textwidth}
     \centering
        \includegraphics[width=1\textwidth]{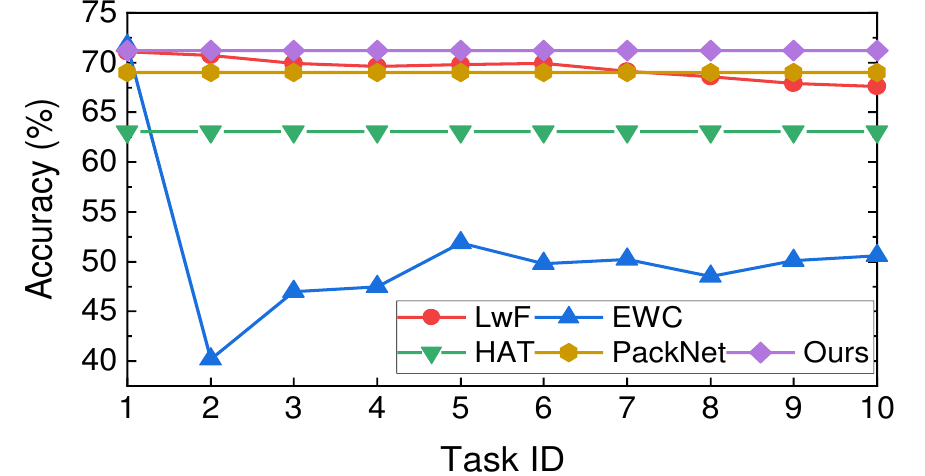} 
        \caption{Task 1 accuracy upon incrementally learning split CIFAR-100 tasks.} 
        \label{fig:longseq}
    \end{minipage}\hfill
    \begin{minipage}{0.47\textwidth}
      \centering
        \includegraphics[width=1\textwidth]{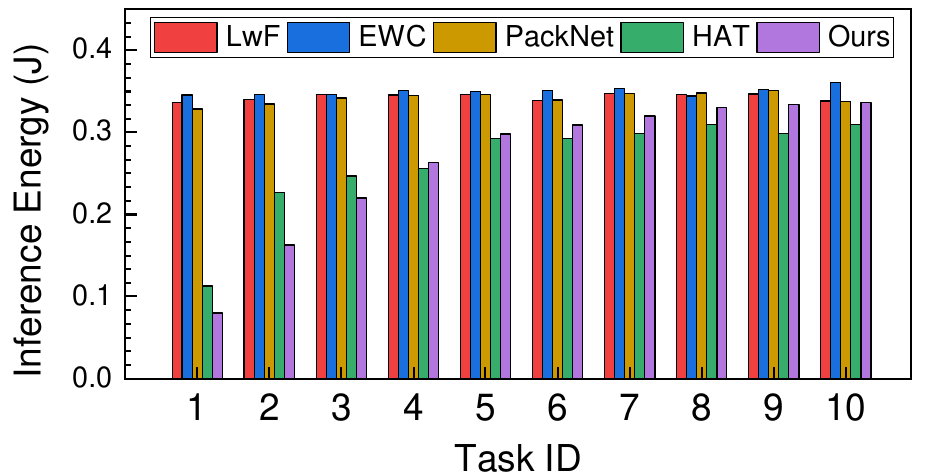} 
        \caption{Comparison of inference energy for learning split CIFAR-100 tasks.}
    \label{fig:energy}
    \end{minipage}
\end{figure}

\textbf{Overcoming Catastrophic Forgetting.} We observe that for longer task sequences such as in split CIFAR-100, the accuracy of Task 1 drops by $21\%$ and $3.5\%$ in EWC and LwF, respectively after the final task is learned incrementally. These methods forget earlier tasks as they learn incrementally, whereas PackNet, HAT and our method preserve the initial performance on each task (Figure \ref{fig:longseq}). Typically, in continual learning algorithms, the original accuracy for tasks degrades  as the number of tasks increase, but all three algorithms overcome this by freezing the relevant parameters. However, due to this freezing of parameters, these methods will eventually hit the resource ceiling as the number of tasks increase. At that point, in our method, the accuracy for each task would have to be traded off with higher compression (lower variance retention thresholds) or the network would have to be grown dynamically as in \cite{den}.

\begin{figure}[t]
\begin{center}
  \includegraphics[width=1\linewidth]{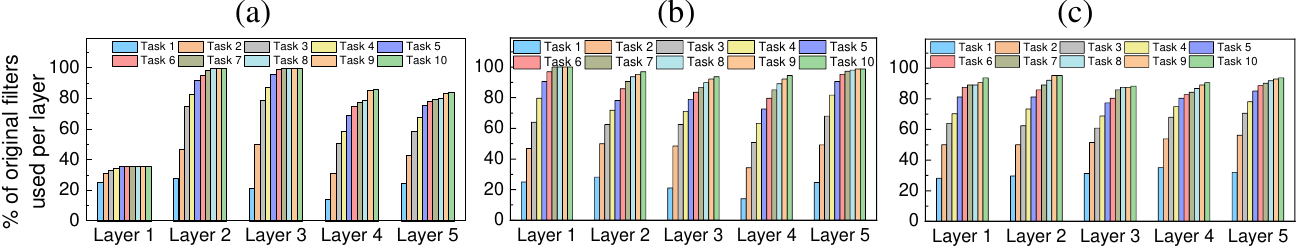}
  \vspace{-4mm}
  \caption{Layer-wise filter utilization for each task for Split CIFAR-100 (a) with and (b) without Projection-Subtraction, using the same variance threshold for pruning. (c) Cumulative distinct filter utilization for HAT.}
  \label{fig:fstat}
\end{center}
\vspace{-15pt}
\end{figure}

\textbf{Sharing Filters Across Tasks}. Filter statistics obtained using SPACE enable us to probe into the learning dynamics, both spatially (across layers) and temporally (across tasks). Spatially, filters learned at the earlier layers are known to be more general, getting more specialized as we go deeper \cite{fergus_viz}. Accordingly, we expect the filters at the earlier layers to be more shared (transferable) across tasks than those learned in the later layers, translating into a higher requirement of additional filters at later layers for each task. Our experiments conform to this expected trend. For example, in Figure \ref{fig:fstat}(a), no new filters were added to layer 1 after Task 5, whereas all the tasks required the addition of new filters at layers 4 and 5. Temporally, as learning progresses, more filters are added to the core, reflecting a growing knowledge base. We expect to see a decreasing requirement of additional filters for later tasks as the knowledge base becomes sufficient. Results in Figure \ref{fig:fstat}(a) comply with this intuition, since earlier tasks require the addition of a larger number of filters to the core than the later tasks. A similar trend is also observed for P-MNIST and CIFAR-10, as shown in Supplementary Figure \ref{fig:sp_fstats}. Supplementary section \ref{S3.3} analyzes core and residual variances, which provides an intuitive picture of the degree of sharing across tasks and layers.

\textbf{Projection-Subtraction Minimizes Resources}. As discussed in Section \ref{sec2}, Projection-Subtraction ensures that we add a minimal set of filters to the Core from the Residual by decoupling them. To quantify the impact, we learn the 10 sequential CIFAR-100 tasks by analyzing the Residual without Projection-Subtraction. We obtain a classification accuracy of $72.82\%$ with $0.92$ fraction of original network parameter utilization. This small accuracy gain of $\sim 0.29\%$ at the cost of $\sim 16.5\%$ extra parameters indicates that Projection-Subtraction induces resource minimality while maintaining classification performance. The resultant filter statistics without Projection-Subtraction are shown in Figure \ref{fig:fstat}(b).~Comparing these statistics with those shown in Figure \ref{fig:fstat}(a) further highlights the representational redundancy removed by the Projection-Subtraction step. 
HAT learns filter significance during training, and we show the resulting statistics from binarizing the learnt masks in Figure \ref{fig:fstat}(c). We note that it looks similar to our statistics without Projection Subtraction, implying that the optimization objective used by HAT for learning masks does not explicitly enforce sharing of filters over tasks. In particular, an analysis of the first layer proves insightful. The first layer consists of 64 kernels of shape $3\times3$, with 3 channels each. This implies that there can be a maximum of 27 independent filters in this layer. Without Projection -Subtraction step, SPACE utilizes all 64 filters. However, once we perform Projection-Subtraction, we show that we utilize a maximum of 23 filters (in the last task), satisfying the statistical maximum and indicating that we are utilizing the Core at each task well. However, across tasks, HAT utilizes up to 60 distinct filters. A re-implementation of HAT in \cite{hatsv} corroborates HAT's tendency to saturate the earlier layers, limiting accuracy in downstream tasks.                

\textbf{Energy Efficiency Analysis}. Table \ref{tab:tab1} contains the average inference energy of all methods, normalized to STL. The slightly better classification performance of PackNet is due to weight level pruning, which is of finer granularity than our filter level pruning. \cite{packnet} points out a drawback of weight level sparsity in the inability to do simultaneous inference or batch-processing, since the data is no longer task-wise separable after the convolution and ReLU layers. Filter-level pruning ensures a separable output, since separate tasks only interfere at channel levels, enabling us to do batch inference. In addition, it translates to higher energy savings in general purpose hardware, owing to the structured nature of the sparsity. In Figure~\ref{fig:energy}, we compare the task-wise inference energy of different algorithms on the split CIFAR-100 dataset, measured on an NVIDIA GeForce GTX1060 GPU with a batch size of $64$ (more details in Supplementary section \ref{S2.3}, other datasets shown in Supplementary Figure \ref{fig:sp_energy}). While STL has 10 separate networks, only one is engaged for classification, since the task hint is provided. Hence the energy is similar to EWC and LwF, since all three utilize the entire dense network. While PackNet has varying levels of unstructured sparsity for different tasks, we notice that it consumes comparable energy to the dense models, highlighting the inefficiency of GPUs for computation with irregularly sparse matrix kernels~\cite{sparse}. On the other hand, both HAT and SPACE add or remove filters for different tasks, which the computing hardware can leverage to provide considerable energy benefits. Since the network trained for Task 1 is the smallest, SPACE achieves $\sim 5$x inference energy reduction compared to methods utilizing the dense network. As further tasks are learned, the network grows in size and the energy consumption scales accordingly. The last task has the largest model, but still consumes $\sim 1.3$x lesser energy. Averaging over all tasks, SPACE consumes about $60\%$ the amount of energy consumed by LwF, EWC or PackNet. HAT also results in structured sparsity, and therefore it too performs better than the previous baselines. However, the average resource utilization of HAT is higher than our method, resulting in up to 20\% more energy consumption (while also being lesser accurate). Overall, as shown in Table~\ref{tab:tab1}, we show comparably or better inference energy efficiency than all methods for all datasets, primarily due to the sharing and minimality enforced by the Projection-Subtraction step.


\vspace{-5pt}

\section{Conclusion}

In this paper, we address the problem of catastrophic forgetting in DNNs in a CL setting, under the constraint that data from older tasks is not shared, and the learning system has a fixed amount of resources. To accommodate multiple sequential tasks, we divide the representational space into a fixed Core space, which contains task specific information over all previously seen tasks, and a Residual space, that the current task learns over. We propose an algorithm that first decouples the Residual from the Core, and then analyzes this Residual to find a minimal set of filters to add to the Core. The remaining parameters are freed up for the next task, and only one retraining iteration is needed per task, allowing us to scale to deeper networks. We consistently outperform EWC, LwF and HAT in terms of accuracy, network size and inference energy efficiency, and overcome the problem of catastrophic forgetting. While PackNet achieves slightly better accuracy than our algorithm, we are able to leverage the structured sparsity much better in hardware, resulting in up to $5$x improvement in energy efficiency at the earlier tasks (up to $\sim40$\% less energy consumption overall) as compared to PackNet. Structured sparsity also enables us to infer mixed batches over different tasks. 
\medskip


\section{Acknowledgement}
This work was supported in part by the Center for Brain Inspired Computing (C-BRIC), one of the
six centers in JUMP, a Semiconductor Research Corporation (SRC) program sponsored by DARPA,
by the Semiconductor Research Corporation, the National Science Foundation, Intel Corporation,
the DoD Vannevar Bush Fellowship, and by the U.S. Army Research Laboratory and the U.K.
Ministry of Defence under Agreement Number W911NF-16-3-0001.
\begingroup
\small 
\bibliographystyle{unsrt}
\bibliography{neurips_2020}

\begin{thebibliography}{10}

\bibitem{resnet}
Kaiming He, Xiangyu Zhang, Shaoqing Ren, and Jian Sun.
\newblock Deep residual learning for image recognition.
\newblock In {\em CVPR}, June 2016.

\bibitem{algo}
Volodymyr Mnih, Koray Kavukcuoglu, David Silver, Andrei~A. Rusu, Joel Veness,
  Marc~G. Bellemare, Alex Graves, Martin~A. Riedmiller, Andreas~K. Fidjeland,
  Georg Ostrovski, Stig Petersen, Charles Beattie, Amir Sadik, Ioannis
  Antonoglou, Helen. King, Dharshan Kumaran, Daan Wierstra, Shane Legg, and
  Demis Hassabis.
\newblock Human-level control through deep reinforcement learning.
\newblock {\em Nature}, 518:529--533, 2015.

\bibitem{cla}
Mark~B. Ring.
\newblock Child: A first step towards continual learning.
\newblock In {\em Learning to Learn}, 1998.

\bibitem{clb}
S.~Thrun and Tom~Michael Mitchell.
\newblock Lifelong robot learning.
\newblock {\em Robotics Auton. Syst.}, 15:25--46, 1995.

\bibitem{rev}
German~Ignacio Parisi, Ronald Kemker, Jose~L. Part, Christopher Kanan, and
  Stefan Wermter.
\newblock Continual lifelong learning with neural networks: A review.
\newblock {\em Neural networks : the official journal of the International
  Neural Network Society}, 113:54--71, 2019.

\bibitem{cl3}
Xilai Li, Yingbo Zhou, Tianfu Wu, Richard Socher, and Caiming Xiong.
\newblock Learn to grow: A continual structure learning framework for
  overcoming catastrophic forgetting.
\newblock {\em ArXiv}, abs/1904.00310, 2019.

\bibitem{owm}
Guanxiong Zeng, Yang Chen, Bo~Cui, and Shan Yu.
\newblock Continual learning of context-dependent processing in neural
  networks.
\newblock {\em Nature Machine Intelligence}, 1:364–372, 2019.

\bibitem{cat1}
Michael Mccloskey and Neil~J. Cohen.
\newblock Catastrophic interference in connectionist networks: {T}he sequential
  learning problem.
\newblock {\em The Psychology of Learning and Motivation}, 24:104--169, 1989.

\bibitem{cat2}
R.~Ratcliff.
\newblock Connectionist models of recognition memory: constraints imposed by
  learning and forgetting functions.
\newblock {\em Psychological review}, 97 2:285--308, 1990.

\bibitem{robbo}
Anthony Robins.
\newblock Catastrophic forgetting, rehearsal and pseudorehearsal.
\newblock {\em Connection Science}, 7:123--146, 1995.

\bibitem{icarl}
Sylvestre-Alvise Rebuffi, Alexander Kolesnikov, Georg Sperl, and Christoph~H.
  Lampert.
\newblock {iCaRL}: Incremental classifier and representation learning.
\newblock In {\em CVPR}, 2016.

\bibitem{gem}
David Lopez-Paz and Marc'Aurelio Ranzato.
\newblock Gradient episodic memory for continuum learning.
\newblock In {\em NIPS}, 2017.

\bibitem{er}
Arslan Chaudhry, M.~Rohrbach, Mohamed Elhoseiny, Thalaiyasingam Ajanthan,
  P.~Dokania, P.~Torr, and Marc'Aurelio Ranzato.
\newblock Continual learning with tiny episodic memories.
\newblock {\em ArXiv}, abs/1902.10486, 2019.

\bibitem{dgr}
Hanul Shin, Jung~Kwon Lee, Jaehong Kim, and Jiwon Kim.
\newblock Continual learning with deep generative replay.
\newblock In {\em NIPS}, 2017.

\bibitem{ewc}
James Kirkpatrick, Razvan Pascanu, Neil~C. Rabinowitz, Joel Veness, Guillaume
  Desjardins, Andrei~A. Rusu, Kieran Milan, John Quan, Tiago Ramalho, Agnieszka
  Grabska-Barwinska, Demis Hassabis, Claudia Clopath, Dharshan Kumaran, and
  Raia Hadsell.
\newblock Overcoming catastrophic forgetting in neural networks.
\newblock {\em Proceedings of the National Academy of Sciences of the United
  States of America}, 114 13:3521--3526, 2016.

\bibitem{si}
Friedemann Zenke, Ben Poole, and Surya Ganguli.
\newblock Continual learning through synaptic intelligence.
\newblock In {\em ICML}, 2017.

\bibitem{mas}
Rahaf Aljundi, Francesca Babiloni, Mohamed Elhoseiny, Marcus Rohrbach, and
  Tinne Tuytelaars.
\newblock {Memory Aware Synapses}: Learning what (not) to forget.
\newblock In {\em ECCV}, 2018.

\bibitem{cl4}
Nicolas~Y. Masse, Gregory~D. Grant, and David~J. Freedman.
\newblock Alleviating catastrophic forgetting using context-dependent gating
  and synaptic stabilization.
\newblock {\em Proceedings of the National Academy of Sciences of the United
  States of America}, 115 44:E10467--E10475, 2018.

\bibitem{pisol}
Rahaf Aljundi, Min Lin, Baptiste Goujaud, and Yoshua Bengio.
\newblock Gradient based sample selection for online continual learning.
\newblock In {\em NeurIPS 2019}, 2019.

\bibitem{packnet}
Arun Mallya and Svetlana Lazebnik.
\newblock {PackNet}: Adding multiple tasks to a single network by iterative
  pruning.
\newblock In {\em CVPR}, June 2018.

\bibitem{pathnet}
Chrisantha Fernando, Dylan Banarse, Charles Blundell, Yori Zwols, David Ha,
  Andrei~A. Rusu, Alexander Pritzel, and Daan Wierstra.
\newblock {PathNet}: Evolution channels gradient descent in super neural
  networks.
\newblock {\em ArXiv}, abs/1701.08734, 2017.

\bibitem{den}
Jaehong Yoon, Eunho Yang, and Sung~Ju Hwang.
\newblock Lifelong learning with dynamically expandable networks.
\newblock {\em ArXiv}, abs/1708.01547, 2017.

\bibitem{progress}
Andrei~A. Rusu, Neil~C. Rabinowitz, Guillaume Desjardins, Hubert Soyer, James
  Kirkpatrick, Koray Kavukcuoglu, Razvan Pascanu, and Raia Hadsell.
\newblock Progressive neural networks.
\newblock {\em ArXiv}, abs/1606.04671, 2016.

\bibitem{pca}
Herv{\'e} Abdi and Lynne~J. Williams.
\newblock {Principal Component Analysis}.
\newblock {\em WIREs Comput. Stat.}, 2(4):433--459, July 2010.

\bibitem{hat}
Joan Serra, Didac Suris, Marius Miron, and Alexandros Karatzoglou.
\newblock Overcoming catastrophic forgetting with hard attention to the task.
\newblock volume~80 of {\em Proceedings of Machine Learning Research}, pages
  4548--4557, Stockholmsmässan, Stockholm Sweden, 10--15 Jul 2018. PMLR.

\bibitem{lwf}
Zhizhong Li and Derek Hoiem.
\newblock Learning without forgetting.
\newblock {\em IEEE Transactions on Pattern Analysis and Machine Intelligence},
  40:2935--2947, 2016.

\bibitem{mega}
Yunhui Guo, Mingrui Liu, Tianbao Yang, and Tajana Rosing.
\newblock Improved schemes for episodic memory-based lifelong learning.
\newblock {\em Advances in Neural Information Processing Systems}, 33, 2020.

\bibitem{oewc}
Jonathan Schwarz, Wojciech Czarnecki, Jelena Luketina, Agnieszka
  Grabska-Barwinska, Yee~Whye Teh, Razvan Pascanu, and Raia Hadsell.
\newblock {Progress \& Compress}: A scalable framework for continual learning.
\newblock In {\em ICML}, 2018.

\bibitem{ucb}
Sayna Ebrahimi, Mohamed Elhoseiny, Trevor Darrell, and Marcus Rohrbach.
\newblock Uncertainty-guided continual learning with bayesian neural networks.
\newblock In {\em International Conference on Learning Representations}, 2020.

\bibitem{cpg}
Ching-Yi Hung, Cheng-Hao Tu, Cheng-En Wu, Chien-Hung Chen, Yi-Ming Chan, and
  Chu-Song Chen.
\newblock Compacting, picking and growing for unforgetting continual learning.
\newblock In {\em Advances in Neural Information Processing Systems}, pages
  13647--13657, 2019.

\bibitem{acl}
Sayna Ebrahimi, Franziska Meier, Roberto Calandra, Trevor Darrell, and Marcus
  Rohrbach.
\newblock Adversarial continual learning.
\newblock {\em arXiv preprint arXiv:2003.09553}, 2020.

\bibitem{apd}
Jaehong Yoon, Saehoon Kim, Eunho Yang, and Sung~Ju Hwang.
\newblock Scalable and order-robust continual learning with additive parameter
  decomposition.
\newblock In {\em International Conference on Learning Representations}, 2020.

\bibitem{cl2}
Deboleena Roy, Priyadarshini Panda, and Kaushik Roy.
\newblock {Tree-CNN}: A hierarchical deep convolutional neural network for
  incremental learning.
\newblock {\em Neural networks : the official journal of the International
  Neural Network Society}, 121:148--160, 2020.

\bibitem{rpsnet}
Jathushan Rajasegaran, Munawar Hayat, Salman Khan, Fahad~Shahbaz Khan, and Ling
  Shao.
\newblock Random path selection for incremental learning.
\newblock {\em Advances in Neural Information Processing Systems}, 2019.

\bibitem{overparam}
Misha Denil, Babak Shakibi, Laurent Dinh, Marc'Aurelio Ranzato, and Nando
  de~Freitas.
\newblock Predicting parameters in deep learning.
\newblock In {\em NIPS}, 2013.

\bibitem{magprune}
Song Han, Jeff Pool, John Tran, and William~J. Dally.
\newblock Learning both weights and connections for efficient neural network.
\newblock In {\em NIPS}, 2015.

\bibitem{dsd}
Song Han, Jeff Pool, Sharan Narang, Huizi Mao, Enhao Gong, Shijian Tang, Erich
  Elsen, Peter Vajda, Manohar Paluri, John Tran, Bryan Catanzaro, and
  William~J. Dally.
\newblock {DSD: Dense-Sparse-Dense} training for deep neural networks.
\newblock In {\em ICLR}, 2016.

\bibitem{sp}
Wei Wen, Chunpeng Wu, Yandan Wang, Yiran Chen, and Hai Li.
\newblock Learning structured sparsity in deep neural networks.
\newblock In {\em NIPS}, 2016.

\bibitem{transformer}
Aayush Ankit, Abhronil Sengupta, and Kaushik Roy.
\newblock {TraNNsformer}: Neural network transformation for memristive crossbar
  based neuromorphic system design.
\newblock {\em 2017 IEEE/ACM International Conference on Computer-Aided Design
  (ICCAD)}, pages 533--540, 2017.

\bibitem{alex}
Alex Krizhevsky, Ilya Sutskever, and Geoffrey~E. Hinton.
\newblock Imagenet classification with deep convolutional neural networks.
\newblock {\em Commun. ACM}, 60:84--90, 2012.

\bibitem{vgg}
Karen Simonyan and Andrew Zisserman.
\newblock Very deep convolutional networks for large-scale image recognition.
\newblock {\em CoRR}, abs/1409.1556, 2015.

\bibitem{isha}
Isha Garg, Priyadarshini Panda, and Kaushik Roy.
\newblock A low effort approach to structured {CNN} design using {PCA}.
\newblock {\em IEEE Access}, 8:1347--1360, 2020.

\bibitem{stang}
Gilbert Strang.
\newblock {\em Linear algebra and its applications}.
\newblock Thomson, Brooks/Cole, Belmont, CA, 2006.

\bibitem{mnist}
Yann LeCun, L{\'e}on Bottou, Yoshua Bengio, and Patrick Haffner.
\newblock Gradient-based learning applied to document recognition.
\newblock {\em Proceedings of the IEEE}, 86(11):2278--2324, Nov 1998.

\bibitem{cifar}
Alex Krizhevsky.
\newblock Learning multiple layers of features from tiny images.
\newblock 2009.

\bibitem{fergus_viz}
Matthew~D. Zeiler and Rob Fergus.
\newblock Visualizing and understanding convolutional networks.
\newblock {\em CoRR}, abs/1311.2901, 2013.

\bibitem{hatsv}
Matthias~De Lange, Rahaf Aljundi, Marc Masana, Sarah Parisot, Xu~Jia,
  A.~Leonardis, Gregory Slabaugh, and T.~Tuytelaars.
\newblock A continual learning survey: Defying forgetting in classification
  tasks.
\newblock {\em arXiv: Computer Vision and Pattern Recognition}, 2020.

\bibitem{sparse}
Maohua Zhu, Tao Zhang, Zhenyu Gu, and Yuan Xie.
\newblock Sparse tensor core: Algorithm and hardware co-design for vector-wise
  sparse neural networks on modern gpus.
\newblock In {\em MICRO}, 2019.

\bibitem{face}
Hongwei Ng and S.~Winkler.
\newblock A data-driven approach to cleaning large face datasets.
\newblock {\em 2014 IEEE International Conference on Image Processing (ICIP)},
  pages 343--347, 2014.

\bibitem{svhn}
Yuval Netzer, T.~Wang, A.~Coates, Alessandro Bissacco, B.~Wu, and A.~Ng.
\newblock Reading digits in natural images with unsupervised feature learning.
\newblock 2011.

\bibitem{fmnist}
H.~Xiao, K.~Rasul, and Roland Vollgraf.
\newblock Fashion-mnist: a novel image dataset for benchmarking machine
  learning algorithms.
\newblock {\em ArXiv}, abs/1708.07747, 2017.

\bibitem{tsign}
Johannes Stallkamp, Marc Schlipsing, Jan Salmen, and Christian Igel.
\newblock The german traffic sign recognition benchmark: A multi-class
  classification competition.
\newblock In {\em In International Joint Conference on Neural Networks}, pages
  1453--1460, 2011.

\bibitem{nmnist}
Yaroslav Bulatov.
\newblock {\em Notmnist dataset}.
\newblock Google (Books/OCR), Tech. Rep.[Online], 2011.

\bibitem{evaluation}
Yen-Chang Hsu, Yen-Cheng Liu, and Zsolt Kira.
\newblock Re-evaluating continual learning scenarios: A categorization and case
  for strong baselines.
\newblock {\em ArXiv}, abs/1810.12488, 2018.

\bibitem{ygal}
Sebastian Farquhar and Yarin Gal.
\newblock Towards robust evaluations of continual learning.
\newblock {\em ArXiv}, abs/1805.09733, 2018.

\bibitem{dc}
Song Han, Huizi Mao, and William~J. Dally.
\newblock Deep compression: Compressing deep neural network with pruning,
  trained quantization and huffman coding.
\newblock In Yoshua Bengio and Yann LeCun, editors, {\em 4th International
  Conference on Learning Representations, {ICLR} 2016, San Juan, Puerto Rico,
  May 2-4, 2016, Conference Track Proceedings}, 2016.

\bibitem{cu1}
Sharan Chetlur, Cliff Woolley, Philippe Vandermersch, Jonathan Cohen, John
  Tran, Bryan Catanzaro, and Evan Shelhamer.
\newblock {cuDNN}: Efficient primitives for deep learning.
\newblock {\em ArXiv}, abs/1410.0759, 2014.

\bibitem{cu2}
Nathan Bell and Michael Garland.
\newblock Efficient sparse matrix-vector multiplication on cuda.
\newblock 2008.

\bibitem{rtf}
Michiel van~der Ven and Andreas~S. Tolias.
\newblock Generative replay with feedback connections as a general strategy for
  continual learning.
\newblock {\em ArXiv}, abs/1809.10635, 2018.

\end{thebibliography}
\endgroup
\clearpage
\newcommand\HRule{\rule{\textwidth}{1pt}}
\HRule \\[0.4cm]
\vspace{-25pt}
\begin{center}
\textbf{\LARGE Supplementary Materials}
\end{center}
\vspace{-10pt}
\HRule \\[0.4cm]
\setcounter{equation}{0}
\setcounter{figure}{0}
\setcounter{table}{0}
\setcounter{page}{1}
\makeatletter
\renewcommand{\theequation}{S\arabic{equation}}
\renewcommand{\thefigure}{S\arabic{figure}}
\renewcommand{\thetable}{S\arabic{table}}
\setcounter{section}{0}
\renewcommand{\thesection}{S\arabic{section}}%

\vspace{-10pt}
\section{Algorithm Details}

\subsection{Projection of the Residual on the Core}\label{S1.1}
As discussed in Section \ref{sec2}, in the \textbf{Projection} step, we construct $A_r^{proj}$ by projecting $A_r$ onto the space described by the (column) basis of $A_f$. Here, $A_r\in \mathbb{R}^{n\times r}$ contains the activations generated by the Residual filters and $A_f\in \mathbb{R}^{n\times f}$ by the frozen Core filters. We apply reduced Singular Value Decomposition (SVD) on $A_f$:
\begin{equation}\label{eq:s_eq1}
    A_f=U_f\Sigma_f V_f^T,
\end{equation}
where only non-zero singular values in $\Sigma_f$ and corresponding singular vectors in $U_f$ and $V_f$ are retained. Thus, we obtain $U_f\in \mathbb{R}^{n\times f}$, whose column vectors are the desired orthonormal basis of the space that we want $A_r$ to project on. The projection is given by:
\begin{equation}\label{eq:s_eq2}
    A_r^{proj}=(U_fU_f^T)A_r.
\end{equation}
This matrix represents the closest approximation of the Residual by the Core or the part of Residual that the Core can reconstruct without any additional filters. Thus the frozen Core explains a part of the variance of the Residual, and the filters needed are far lesser than if we did not perform this Projection. 

The $U_f$ matrix is computed by decomposing $A_fA_f^T$, a matrix of dimensions $n\times n$. Since in our experiments $n>>f$, it is desirable to avoid decomposition of $A_fA_f^T$ to compute $U_f$. Therefore, we compute $\Sigma_f$ and $V_f$ by decomposing  $A_f^TA_f$, and then compute $U_f$ from Equation \ref{eq:eq1}. Since $A_f^TA_f$ is a $f\times f$ matrix (reasonably small in size in our experiments), the computation of $U_f$ for projection does not add much computational overhead.

\subsection{Algorithm Pseudo Code}\label{S1.2}
In Algorithm \ref{alg:taskT}, we outline the steps for learning any sequential task in the form of a Pseudo code. The pseudocode utilizes the functions outlined in Algorithm \ref{alg:PCA_net}, where we show the network compression steps using PCA for learning the first task, and in Algorithm \ref{alg:PCA_sel}, which shows the steps for Projection-Subtraction followed by PCA. 

\begin{algorithm}
   \caption{Learning a Sequential Task}
   \label{alg:taskT}
\begin{algorithmic}[1]
   \STATE {\bfseries Input:} Current dataset, $D_t=\{(X_i,y_i)\}_{i=1}^n$; Task ID, $t$; network model, $f_{\theta}^{t-1}$; dict of core filters, $F_l$; variance threshold, $v_{th}$ 
  \STATE {\bfseries Output:} updated network model, $f_{\theta}^{t}$; updated dict of core filters, $F_l$ 
  \STATE
   \STATE
// train the network on current dataset while keeping the core filters (if any) frozen
    \STATE $f_{\theta}^{t}$ $\gets$ train ($f_{\theta}^{t-1}$, $D_t$, $F_l$ ) 
    \STATE
    \STATE // get the number of filters per layer for the current task
    \IF{t == 1} 
   \STATE current\_filters $\gets$ \textit{\textbf{Network\_Compression\_PCA}}($f_{\theta}^{t}$, $D_t$, $v_{th}$) //(Algorithm \ref{alg:PCA_net})
    \ELSE
        \STATE current\_filters $\gets$ \textit{\textbf{Projection\_Subtraction\_PCA}} ($f_{\theta}^{t}$, $F_l[t-1]$, $D_t$, $v_{th}$) // (Algorithm \ref{alg:PCA_sel})
    \ENDIF

    \STATE $F_l[t] \gets$ current\_filters
    \STATE
    \STATE // prune the unimportant portion after PCA based on current filter list
    \STATE $f_{\theta}^{t}$ $\gets$prune\_model ($f_{\theta}^{t}$, $F_l[t]$) 
    \STATE
    \STATE // retrain the network on current dataset while keeping the core filters frozen
    \STATE $f_{\theta}^{t}$ $\gets$ retrain ($f_{\theta}^{t}$, $D_t$, $F_l$ ) 
\end{algorithmic}
\end{algorithm}
   
\begin{algorithm}
   \caption{Network Compression PCA Algorithm}
   \label{alg:PCA_net}
\begin{algorithmic}[1]
   \STATE \textbf{function} \textit{Network\_Compression\_PCA} ($f_{\theta}^{t}$, $D_t$, $v_{th}$)
   \STATE
   \STATE // Data is forward propagated trough the model and pre-ReLU activations in each layer are stored 
   \STATE activation\_list $\gets$ collect\_activation ($f_{\theta}^{t}$, $D_t$) 
  \STATE current\_filters $\gets$ zeros($L-1$) \space // $L$ is the number of layers in the network
\STATE
  \FOR{$l \leftarrow 1$ {\bfseries to} $L-1$}   
  \STATE // for each layer except the classifier 
  \STATE $n_0$ $\gets$ activation\_list[$l$].size(2) \space\space// \# original filters in layer $l$
  \STATE $A$ $\gets$ activation\_list[$l$].reshape(-1,$n_0$) \space\space // $A$ is $n\times n_0$ matrix (see Section \ref{sec1})  
  \STATE $A$ $\gets$ normalize (A) \space\space //column mean normalization
  \STATE $n$ $\gets$ A.size(1)\space\space// \# samples
  \STATE
  \STATE //Determination of number of filters using PCA
  \STATE pca $\gets$ \textbf{PCA} ($A$)
  \STATE var\_ratio $\gets$ pca.explained\_variance\_ratio\_  \space\space// an $n_0$ element array
  \STATE $x' \gets$ $0$ 
    \STATE current\_filters[$l$] $\gets 0$
    \FOR{$i \leftarrow 1$ {\bfseries to} length(var\_ratio)}
    \IF{$x'$<$v_{th}$} 
     \STATE // variance accumulation until threshold
    \STATE $x' \gets x'$ + var\_ratio[$i$]
    \STATE current\_filters[$l$] $\gets$current\_filters[$l$]+ 1
    \ENDIF
    \ENDFOR
\ENDFOR
\STATE \textbf{return} current\_filters
\STATE \textbf{end function}
\end{algorithmic}
\end{algorithm}

\begin{algorithm}
   \caption{Projection-Subtraction-PCA Algorithm}
   \label{alg:PCA_sel}
\begin{algorithmic}[1]
   \STATE \textbf{function} \textit{Projection\_Subtraction\_PCA} ($f_{\theta}^{t}$, core\_filter\_list, $D_t$, $v_{th}$)
   \STATE
   \STATE // Data is forward propagated trough the model and pre-ReLU activations in each layer are stored 
   \STATE activation\_list $\gets$ collect\_activation ($f_{\theta}^{t}$, $D_t$) 
  \STATE current\_filters $\gets$ zeros ($L-1$) \space //$L$ is the number of layers in the network 
\STATE
  \FOR{$l \leftarrow 1$ {\bfseries to} $L-1$}   
  \STATE // for each layer except the classifier 
  \STATE $f$ $\gets$ core\_filter\_list[$l$] \space\space// \# core filters
  \STATE $n_0$ $\gets$ activation\_list[$l$].size(2) \space\space// \# original filters in layer $l$
    \STATE $r$ $\gets$ ($n_0-f$) \space\space // \# residual filters
  
  \STATE
   \STATE //Activation partition and variances
  \STATE $A$ $\gets$ activation\_list[$l$].reshape(-1,$n_0$) \space\space // $A$ is $n\times n_0$ matrix (see Section \ref{sec1})  
  \STATE $A$ $\gets$ normalize (A) \space\space //column mean normalization
  \STATE $n$ $\gets$ A.size(1)\space\space// \# samples
  \STATE $A_f$ $\gets$ $A[:,0:f]$   
  \STATE $A_r$ $\gets$ $A[:,f:end]$  
  \STATE $v_f$ $\gets$ $\frac{1}{n-1}\text{tr}(A_f^TA_f)$ \space\space //core variance
  \STATE $v_r$ $\gets$ $\frac{1}{n-1}\text{tr}(A_r^TA_r)$ \space\space//residual variance
  \STATE
  \STATE //Projection-Subtraction-PCA
  \STATE $A_r^{proj}$ $\gets$ \textbf{Project} ($A_f,A_r$) \space\space // (see Section \ref{sec2} \& Equation \ref{eq:eq1})
    \STATE $v_r^{proj}$ $\gets$ $\frac{1}{n-1}\text{tr}((A_r^{proj})^T A_r^{proj})$ \space\space//residual variance explained by core 
  \STATE  $A_r^{new}$ $\gets$ $A_r-A_r^{proj}$ \space\space // \textbf{Subtraction} (see Section 3.3 \&  Equation \ref{eq:eq2})
  \STATE pca $\gets$ \textbf{PCA} ($A_r^{new}$)
  \STATE
  \STATE //Determination of filter number   
  \STATE var\_ratio $\gets$ pca.explained\_variance\_ /$v_r$ \space\space // an $r$ element array
  \STATE $x' \gets$ $(v_r^{proj}/v_r)$ 
    \STATE current\_filters[$l$] $\gets f$
    \FOR{$i \leftarrow 1$ {\bfseries to} length(var\_ratio)}
    \IF{$x'$<$v_{th}$}
    \STATE $x' \gets x'$ + var\_ratio[$i$]
    \STATE current\_filters[$l$] $\gets$current\_filters[$l$]+ 1
    \ENDIF
    \ENDFOR
    \ENDFOR
\STATE \textbf{return} current\_filters
\STATE \textbf{end function}
\end{algorithmic}
\end{algorithm}

\section{Configurations}

\subsection{Dataset Statistics, Architecture and Training Configurations}\label{S2.1}
\textbf{8-Datasets Statistics.} The summary of 8-dataset~\citep{hat} is shown in Table~\ref{tab:sp_tab0}. Similar to~\cite{hat}, we resize the images to $32\times 32\times 3$ if necessary. For datasets with monochromatic images, we replicate the image across all RGB channels.

\renewcommand{\arraystretch}{1.1}
\begin{table}[h]
\centering
\small
\begin{tabular}{@{}lcccc@{}}\toprule
&  \multicolumn{1}{c}{\#~Classes} &  \multicolumn{1}{c}{Train} & \multicolumn{1}{c}{Validation} & \multicolumn{1}{c}{Test}\\
\midrule
CIFAR-10~\citep{cifar}     & 10  &   45,000  & 5,000  & 10,000  \\
CIFAR-100~\citep{cifar}    & 100 &   45,000  & 5,000  & 10,000 \\
FaceScrab~\citep{face}     & 100 & 18,540 & 2,060 & 2,289\\
MNIST~\citep{mnist} & 10 & 54,000 & 6,000 & 10,000\\
SVHN~\citep{svhn} & 10 & 65,931 & 7,326 & 
 26,032\\
 Fashion-MNIST~\citep{fmnist} & 10 & 54,000 & 6,000 & 10,000\\
 Traffic-Signs~\citep{tsign} & 43 & 35,288 & 3,921 & 12,630\\
 Not-MNIST~\citep{nmnist} &10& 15,168 & 1,685 & 1,873\\
\bottomrule

\end{tabular}
\caption{8-Datasets Statistics.}
\label{tab:sp_tab0}
\end{table}

\textbf{Network Architecture.} For CIFAR-10 and CIFAR-100, we use convolutional neural network (CNN) architectures with five convolutional layers followed by a classifier (\texttt{conv1\--conv2\--pool\--dropout\--conv3\--conv4\--pool\--dropout\--conv5\--pool\--classifier}). For the CIFAR-10 experiments, we use $32-32-64-64-128$ filters in the five convolutional layers whereas for CIFAR-100 we double them to $64-64-128-128-256$ filter. In both experiments, filter kernels for the first four layers are $3\times 3$ while the fifth layer has a kernel size of $2\times 2$. For P-MNIST, we use a Multi Layer Perceptron (MLP) with two hidden layers, having 1000 neurons each, followed by a classifier layer. For 8-dataset tasks, we use a reduced ResNet-18, similar to~\cite{gem}. This network consists of a front convolutional layer (\texttt{conv1}) followed by 4 residual blocks (\texttt{conv2\_x, conv3\_x, conv4\_x, conv5\_x}) each having four convolutional layers (see Table 1 in~\cite{resnet}) followed by a classifier layer. We use 40 as the base number of filters.

\textbf{Training and Test Settings.} All images are pre-processed with zero padding to have dimensions of $32\times 32$ pixels and then normalized to have a mean of 0 and a standard deviation of 1. For all models and algorithms, we use Stochastic Gradient Descent (SGD) with momentum ($0.9$) as the optimizer. The Split CIFAR-10 and CIFAR-100 experiments with CNNs use a dropout of $0.15$ whereas P-MNIST experiments with MLP do not use dropout. For 8-Dataset experiments we use a batch size of $64$ and in all the other experiments we use batch size of $128$. The optimizer is reset after each task in all the experiments. Training epochs, retraining epochs and learning rate schedules for different experiments with P-MNIST, CIFAR-10 and CIFAR-100 datasets (for our algorithm) are listed in Table \ref{tab:sp_tab1}. For each 8-Dataset task, we train (retrain) the network with initial learning rate of $0.01$, and (following HAT~\citep{hat}) decay it by a factor of $3$ if there is no improvement in the validation loss for 5 consecutive epochs. We stop training if we reach a learning rate lower than $10^{-4}$ or we have iterated over $150~(100)$ epochs. In our method, we retrain the network once per task to mitigate any loss in accuracy at the end of the projection-subtraction-PCA steps. We use $1000$ random training examples from the current dataset for activation collection as per~\cite{isha}. We train and test all the models and algorithms in the multi-headed setting~\citep{evaluation,ygal}, where a new classifier is added for each new task, and a task hint is provided during inference. Using that task hint, we activate the relevant portion of the network for inference.

\begin{table}[t] 
\centering 
\begin{tabular}{l c c c c c c c} 
\toprule 
&\multicolumn{5}{c}{Training}
&\multicolumn{1}{c}{Retraining}\\
\cmidrule(l){3-5} 
\cmidrule(l){6-8} 
\textbf{Dataset} & Task ID &  epoch & lr & decay  &  epoch & lr & decay\\ 
\midrule 
\midrule 
P-MNIST & T1 - T7	& 15&	0.01&	6,13&	45&	0.001&	38
 \\ 
 &T8 - T10&	15&	0.01&	6,13&	60&	0.001&	51
 \\ 

\midrule 
CIFAR-10 & T1-T2&	40&	0.01&	25,35&	55&	0.01&	11,49
 \\ 
 & T3-T5&	60	&0.001	&54	&85&	0.01&	8,76
 \\ 

\midrule 
CIFAR-100 & T1 - T2 & 80 & 0.01 & 25,65 & 120 & 0.01 & 42,108\\ 
& T3 - T10 & 100 & 0.001 & 90 & 150 & 0.01 & 52,135\\

\bottomrule 
\end{tabular}
\vspace{10pt}
\caption{ Training and retraining settings used for different experiments in the paper (for our algorithm). Here, `lr' denotes initial learning rate for training/retraining whereas `decay' denotes the epoch at which learning rate is decayed by a factor of 10. } 
\label{tab:sp_tab1} 
\end{table}

\subsection{Tuning the Variance Threshold}\label{S2.2}
For P-MNIST, $99.9\%$ variance retention in the first FC layer and $99.5\%$ in the second FC layer were required for good accuracy. For split CIFAR-10, which has two classes per tasks, we could achieve high performance with $x=99.5$ for the first layer and $x=95$ for the other layers. For the 10-class higher complexity tasks of split CIFAR-100, we experimented with $x=99.9$ for the first layer and $x=99$ for the other layers for all tasks. For 8-Dataset experiment, we used $x=99.5$ in the first layer (\texttt{conv1}) and $x=\{95, 93, 92, 91\}$ in the four subsequent residual blocks (\{\texttt{conv2\_x,conv3\_x,conv4\_x,conv5\_x}\}) of the ResNet. Empirically, such block-wise decreasing variance threshold enabled better accuracy-resource trade-off under distribution shift of $8$ diverse datasets.

 \begin{figure}[h]
\begin{center}
  \includegraphics[width=0.9\linewidth]{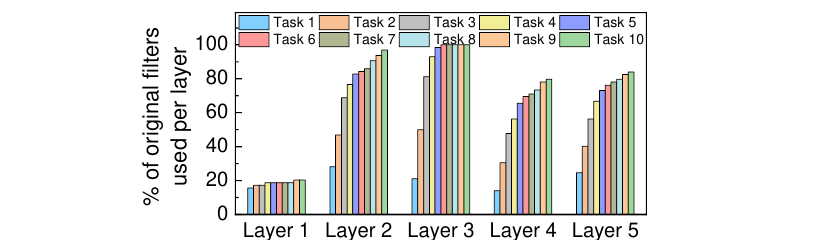}    
\end{center}
  \caption{Filter utilization statistics for 10 sequential CIFAR-100 tasks with uniform variance threshold ($99\%$) for all layer with projection-subtraction. Average accuracy after learning all task was found to be $71.38\%$ with $0.73$ fraction of original network parameter utilization.}
\label{fig:stats}
\end{figure} 

As discussed in Section 5, selection of variance threshold depends on the complexity of the dataset and tasks. This threshold tuning can also be seen as a knob for accuracy-resource trade-off \cite{isha}. In continual learning, setting very high $x$ can potentially allocate large resources (parameters) for one task leaving a lesser-than-sufficient space for the remaining tasks. Judicious choice of $x$ can thus be beneficial in resource constrained environments. Empirically, we found that using higher variance at layer 1 than the other layers leads to better classification performances with small increase in resources. To validate this, we learn 10 sequential CIFAR-100 tasks with uniform variance threshold ($99\%$) for all layer with projection-subtraction. Average accuracy after learning all task is found to be $71.38\%$ with $0.73$ fraction of original network parameter utilization. In Figure \ref{fig:stats}, we show task wise filter utilization statistics. Very low filter utilization in layer 1 motivate us to increase $x$ for layer 1. Thus we choose $x=99.9$ for layer 1 in the CIFAR-100 experiment reported in the main text which yields more than $1\%$ gain in classification accuracy with small increase in filter utilization. Likewise, in the P-MNIST, CIFAR-10 and 8-Datasets experiments, we choose higher variance in layer 1 than the other layers.

 \subsection{Energy Measurements}\label{S2.3}
\begin{figure}[h]
\begin{center}
  \includegraphics[width=0.9\linewidth]{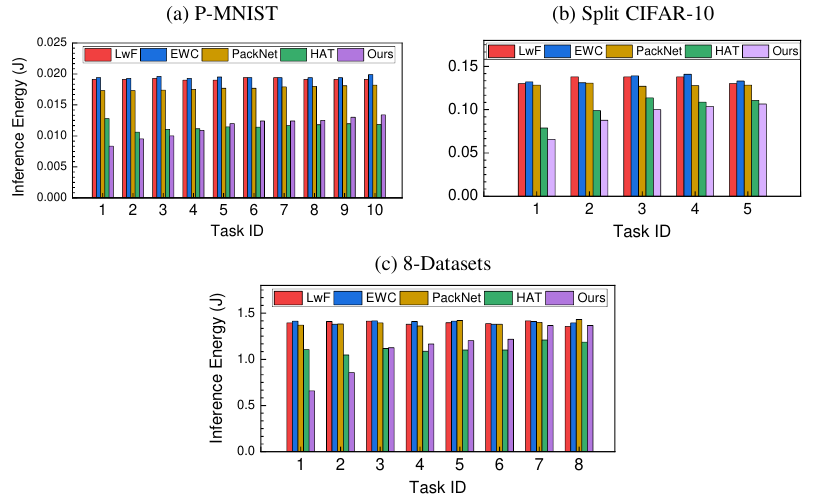}    
\end{center}
  \caption{Comparison of inference energy for learning (a) P-MNIST, (b) Split CIFAR-10 and (c) 8-Datasets tasks.}
\label{fig:sp_energy}
\end{figure}

GPU Inference Energy is computed for a batch of 64 by multiplying power obtained from \texttt{nvidia-smi} command with the computation time as per Table 8,9 of Deep Compression~\citep{dc}. An average of 1000 runs is reported with necessary steps taken to ensure the fairness of comparison with other methods. For instance, 
in PackNet, we noticed that the inclusion of binary mask application in the measurements leads to a $\sim70\%$ increase in energy consumption, indicating an absence of a native binary mask implementation in cuDNN~\cite{cu1} or CUSPARSE~\cite{cu2} libraries. Therefore, we exclude the mask application energy, for a fair comparison. Comparison of inference energy for learning P-MNIST, Split CIFAR-10 and 8-Dataset tasks are given in Figure \ref{fig:sp_energy}(a), \ref{fig:sp_energy}(b) and \ref{fig:sp_energy}(c) respectively. For STL, even though the network size is 10x (for instance for 10 tasks) since we have separate networks learning separate tasks, the inference energy (reported in Table 1 in the main manuscript) is almost the same as a single dense network. This is because we infer with a task hint, that tells us which of the 10 networks to activate, and only that network is used for inference. The numbers reported for STL are averaged over all the ten networks used over the dataset.

\subsection{Baseline Implementations and Hyperparameters}\label{S2.4}
LwF and PackNet are implemented from the official implementation of PackNet~\cite{packnet}, EWC is adopted from the implementation provided by~\cite{rtf}, whereas we implement HAT from its official implementation~\cite{hat}. For EWC, regularization coefficient is set to $1000$ in 8-Dataset experiments and $100$ in all the other experiments, optimized by searching over \{$0.1,1,10,100,1000$\} for best accuracy after all tasks. In LwF (as suggested in~\cite{lwf}), temperature parameter ($T$) is set to 2, regularization strength ($\lambda_0$) is set to 1 and for each new task only the classifier is tuned in the first 5 epochs. In PackNet, compression-ratio ($\%$ pruning) per layer per task is a hyperparameter which needs to be preassigned before training. For PackNet baselines, networks are pruned by 91\% (for P-MNIST), 78\% (for Split CIFAR-10), 85\% (for Split CIFAR-100) and 75\% (for 8-Datasets) after each new task is added. These pruning percentages are chosen to match the final network size from PackNet to our method for a fair comparison of accuracy. HAT has two hyperparameters: stability parameter, $s_{max}$ and compressibility parameter, $c$. We set $c$ to $0.05, 0.1, 0.25$ and $0.5$ for CIFAR-10, CIFAR-100, 8-Datasets and P-MNIST tasks respectively, after searching over $c\in [0.05,0.75]$. We set $s_{max}=400$ for all datasets, optimized by searching over $s_{max}\in [25,800]$ to obtain best average classification performance for the given network.

\section{Additional Analysis}

\subsection{Comparison with Other Methods}\label{here}

We additionally compare our method with UCB~\cite{ucb}. For 10 sequential P-MNIST tasks a small MLP network (0.1M parameters), similar to UCB, is chosen. Whereas for 8-Dataset tasks, similar to UCB, we use ResNet-18 architecture. Results are shown in Table~\ref{tab:tab_ucb}, where UCB results are taken from~\cite{ucb}. For both datasets our method outperforms UCB. 
\begin{table}[t] 
\small
\centering 
\begin{tabular}[t]{l  c c  c c c}

    \toprule 
    &\multicolumn{2}{c}{P-MNIST}
    &\multicolumn{3}{c}{8-Dataset}\\
    \cmidrule(l){2-3} 
    \cmidrule(l){5-6} 
      & UCB$\dagger$  & SPACE(ours)  & &   UCB$\dagger$  & SPACE(ours)\\ 
    \midrule 
    \midrule
    Accuracy(\%) 	&	91.44&	\textbf{93.45}&  &	84.04 &	\textbf{85.48}
     \\ 

    \bottomrule 
    \end{tabular}
\vspace{10pt}
\caption{ Comparison with UCB. ($\dagger$) indicates results reported from UCB. } 
\label{tab:tab_ucb} 
\end{table}

\subsection{Network Utilization Statistics }

\begin{figure}[h]
\begin{center}
  \includegraphics[width=0.9\linewidth]{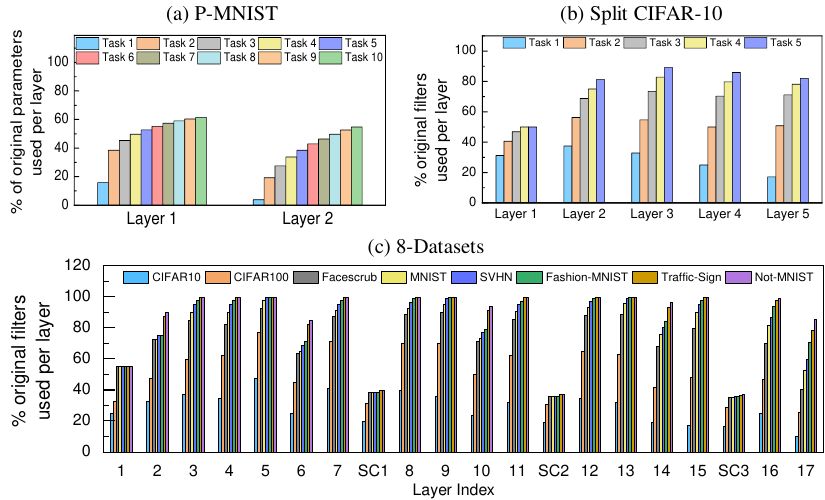}    
\end{center}
  \caption{Layer-wise (a) network parameter utilization for P-MNIST for each tasks. Filter utilization for (b) Split CIFAR-10 and (c) 8-datasets for each tasks. SC denotes parameterized short-cut connections in ResNet.}
\label{fig:sp_fstats}
\end{figure}

Network resource utilization per layer for P-MNIST, CIFAR-10 and 8-Datasets tasks is shown in  Figure~\ref{fig:sp_fstats}. 

\begin{figure}[!h]
\begin{center}
  \includegraphics[width=1.0\linewidth]{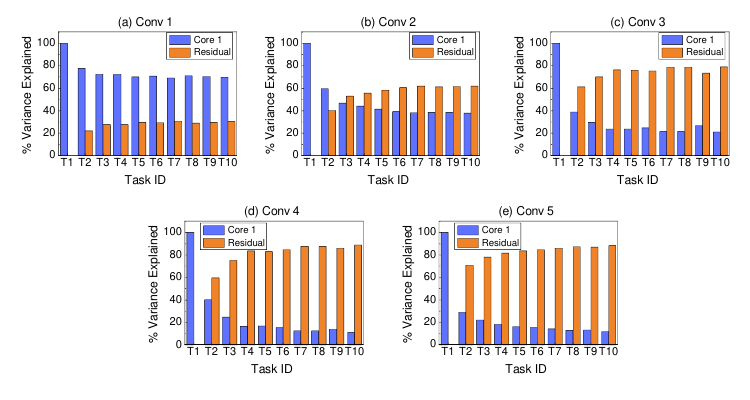}    
\end{center}
  \caption{Plots of variance explained by the Core Space 1 (core filters of Task 1) and residual filters required for a particular CIFAR-100 task at different layers. For each tasks, in early layers, (a) Conv 1 and (b) Conv 2, core space 1 explains higher variance whereas in the later layers, (d) Conv 4 and (e) Conv 5, variance explained is very low. This explains in the initial layer features (or filters) are more shared across tasks while later layers learns task specific features.}
\label{fig:core_res}
\end{figure}

\subsection{On Filter Transferability Across Tasks}\label{S3.3} 
We refer to filters as transferable (shared) across tasks if the filters learned and frozen at one task contribute significantly to the variance explained at the later tasks. Thus, we can treat the `variance explained' by some spaces (core and residual) for a particular task as a metric of utilization of that space for that task. In Figure \ref{fig:core_res}, we show the variance explained by 2 fixed spaces on the Y axis: (i) the compressed Core space learned at the end of Task 1 (Core 1), and (ii) the remaining residual space of that corresponding task, plotted against the task being learned. There are 5 graphs, one for each layer of the network learned on sequential CIFAR-100 tasks. It can be seen that, the Core 1 explains a significant amount of variance when Task 10 is being learned at Layer 1, but not at Layer 5. At Layer 1, Task 1's core filters are general features like edge and color blob detectors and contribute to explaining variances at all tasks. However, as explained in Section 5, filters get more specialized as we go deeper, and we expect the Core space of Task 1 to play a lesser significant role for the later tasks. We can observe this trend for all the five layers, as the core 1 contribution to variance explanation for the later task decreases from Conv 1 to Conv 5 layer. Such trend explains in the initial layer features are more shared across tasks while later layers learns task specific features.

\end{document}